\pdfoutput=1
\documentclass[11pt,letterpaper]{article}
\usepackage{naaclhlt2015}
\usepackage{times}
\usepackage{latexsym}

\usepackage{epstopdf}
\usepackage[ruled]{algorithm2e}
\usepackage{graphicx}
\usepackage{multirow}
\usepackage{url}
\usepackage{subcaption}
\usepackage{amsmath,amsthm,amssymb}

\newcommand{\cnn}{CNN}

\newcommand{\scnnpfx}{seq}

\newcommand{\scnn}{seq-CNN}
\newcommand{\sscnn}{seq2-CNN}
\newcommand{\ssbcnn}{seq2-bow$n$-CNN}

\newcommand{\sconv}{seq-convolution}

\newcommand{\bcnn}{bow-CNN}
\newcommand{\bconv}{bow-convolution}

\newcommand{\bow}{{bow}}
\newcommand{\bowone}{{bow1}}
\newcommand{\bowtwo}{{bow2}}
\newcommand{\bowthree}{{bow3}}
\newcommand{\bongram}{bag-of-$n$-gram}
\newcommand{\bongrams}{bag-of-$n$-grams}

\newcommand{\bx}{{\mathbf x}}
\newcommand{\bz}{{\mathbf z}}

\newcommand{\psz}{p}
\newcommand{\doc}{D}
\newcommand{\voc}{V}
\newcommand{\vsz}{|\voc|}
\newcommand{\wrd}{w}
\newcommand{\dsz}{|\doc|}

\newcommand{\Activ}{\boldsymbol{\sigma}}
\newcommand{\activ}{\sigma}
\newcommand{\ip}[2]{{#1} \cdot {#2}} 

\newcommand{\Wei}{{\mathbf W}}
\newcommand{\Bias}{{\mathbf b}}
\newcommand{\region}{{\mathbf r}} 
\newcommand{\iL}{\ell} 

\newcommand{\Elec}{Elec} 

\newcommand{\bEqsz}{\begin{small}}
\newcommand{\eEqsz}{\end{small}}

\newcommand{\nbw}{NB-LM}

\makeatletter
\newcommand\tightpara{\@startsection{paragraph}{4}{\z@}{1ex plus
   0ex minus 0.2ex}{-1em}{\normalsize\bf}}
\newcommand\normalpara{\@startsection{paragraph}{4}{\z@}{1.5ex plus
   0.5ex minus .2ex}{-1em}{\normalsize\bf}}
\makeatother

\title{Effective Use of Word Order for Text Categorization \\ with Convolutional Neural Networks\thanks{
To appear in NAACL HLT 2015. 
}}
\renewcommand\footnotemark{}
\author{Rie Johnson\\
	    RJ Research Consulting\\
	    Tarrytown, NY, USA\\
	    {\tt riejohnson@gmail.com}
	  \And
	Tong Zhang\\
    Baidu Inc., Beijing, China \\    
    Rutgers University, Piscataway, NJ, USA \\
  {\tt tzhang@stat.rutgers.edu}
}
\date{}
\begin{document}
\maketitle
\begin{abstract}
Convolutional neural network (\cnn) is a neural network that can make use of 
the internal structure of data such as the 2D structure of image data.  
This paper studies \cnn\ on text categorization to exploit
the 1D structure (namely, word order) of text data for accurate prediction.  
Instead of using low-dimensional word vectors as input as is often done, 
we directly apply \cnn\ to high-dimensional text data, which leads to 
directly learning embedding of small text regions for use in classification.  
In addition to a straightforward adaptation of \cnn\ from image to text, 
a simple but new variation which employs bag-of-word conversion in the 
convolution layer is proposed.  
An extension to combine multiple convolution layers is also explored for higher 
accuracy.    
The experiments demonstrate the effectiveness of our approach in comparison with 
state-of-the-art methods.

\end{abstract}

\section{Introduction}

Text categorization is the task of automatically assigning pre-defined categories to 
documents written in natural languages.  
Several types of text categorization have been studied, each of which deals with 
different types of documents and categories, 
such as topic categorization to detect discussed topics (e.g., sports, politics), 
spam detection \cite{SDHH98}, 
and sentiment classification \cite{PLV02,PL08,MDPHNP11} 
to determine 
the sentiment typically in product or movie reviews.  
A standard approach to text categorization 
is to represent documents by {\em bag-of-word vectors}, 
namely, vectors that indicate which words appear in the documents but do not preserve
word order, and use classification models such as 
SVM. 

It has been noted that loss of word order caused by bag-of-word vectors 
({\em \bow\ vectors}) is particularly problematic on sentiment classification.  
A simple remedy is to use word bi-grams in addition to uni-grams 
\cite{BDP07,GBB11,WM12}. 
However, 
use of word $n$-grams with $n>1$ on text categorization in general is not always 
effective; e.g.,   
on topic categorization, simply adding phrases or $n$-grams 
is not effective (see, e.g., references in \cite{TWL02}).  

To benefit from word order on text categorization, 
we take a different approach, 
which employs {\em convolutional neural networks (\cnn)}
\cite{LeCun+etal98}. 
\cnn\ is a neural network that can make use of the internal structure of data 
such as the {\em 2D structure} of image data through convolution layers, 
where each computation unit responds to a small region of input data 
(e.g., a small square of a large image).  We apply \cnn\ to text 
categorization to make use of the {\em 1D structure} (word order) 
of document data so that each unit in the convolution layer responds to 
a small region of a document (a sequence of words).

\cnn\ has been very successful on image classification; see e.g., 
the winning solutions of ImageNet Large Scale Visual Recognition Challenge 
 \cite{imagenetNips12,Szegedy+etal14,Russakovsky+etal14}.  

On text, 
since the work on token-level applications (e.g., POS tagging) by 
\newcite{nnnlpJMLR11}, 
\cnn\ has been used in systems for 
entity search,  
sentence modeling, 
word embedding learning, 
product feature mining, 
and so on 
\cite{XS13,Gao+etal14,Shen+etal14,KGB14,XLLZ14,Tang+etal14,WCA14,Kim14}.  
Notably, in many of these \cnn\ studies on text, 
the first layer of the network converts words in sentences 
to {\em word vectors} by table lookup.  
The word vectors are either trained as part of \cnn\ training, or 
fixed to those learned by some other method 
(e.g., word2vec \cite{wvecNips13}) from an additional large corpus.  
The latter is a form of semi-supervised learning, which we study elsewhere. 
We are interested in the effectiveness of \cnn\ itself 
{\em without aid of additional resources}; therefore, 
word vectors should be trained as part of network training 
if word vector lookup is to be done.  
%

A question arises, however, whether word vector lookup in a purely supervised setting 
is really useful for text categorization.  
The essence of convolution layers is to {\em convert text regions 
of a fixed size (e.g., ``am so happy'' with size 3) to feature vectors}, as described later. 
In that sense, a word vector learning layer is a special (and unusual) case of convolution layer with 
region size one.  
Why is size one appropriate if bi-grams are more discriminating than uni-grams? 
%
Hence, we take a different approach.  
We {\em directly apply \cnn\ to high-dimensional one-hot vectors};  
i.e., we {\em directly} learn {\em embedding}\footnote{
  We use the term `embedding' loosely to mean a structure-preserving function, 
  in particular, a function that generates low-dimensional features 
  that preserve the predictive structure. 
} of text regions without going through word embedding learning.  
This approach is made possible by solving the computational issue\footnote{
  \cnn\ implemented for image would not handle sparse data efficiently, and 
  without efficient handling of sparse data, 
  convolution over high-dimensional one-hot vectors would be computationally infeasible. 
}
through efficient handling of high-dimensional sparse data 
on GPU, and 
it turned out to have the merits of 
improving accuracy 
with fast training/prediction 
and simplifying the system (fewer hyper-parameters to tune).  
Our 
\cnn\ code for text is publicly available on the internet\footnote{
  \url{riejohnson.com/cnn_download.html}
}.  

We 
study the effectiveness of \cnn\
on text categorization and explain why \cnn\ is suitable for the task.  
Two types of \cnn\ are tested: 
{\em \scnn} is a straightforward adaptation of \cnn\ from image to text, and 
{\em \bcnn} is a simple but new variation of \cnn\ that employs bag-of-word conversion 
in the convolution layer.  
The experiments show that \scnn\ outperforms \bcnn\ on sentiment classification, 
vice versa on topic classification, 
and the winner generally outperforms the conventional \bongram\ vector-based methods, as well as 
previous \cnn\ models for text which are more complex. 
%
In particular, to our knowledge, this is the first work that has successfully used 
word order to 
improve topic classification performance.  
%
A simple extension that combines multiple convolution layers (thus 
combining multiple types of text region embedding) 
leads to further improvement.  
Through empirical analysis, we will show that 
\cnn\ can make effective use of high-order $n$-grams when conventional methods fail.  

\section{\cnn\ for document classification}
\label{sec:method}
 
We first review \cnn\ applied to image data and then 
discuss the application of \cnn\ to document classification tasks 
to introduce \scnn\ and \bcnn.  

\begin{figure}
\centering
\includegraphics[width=2.6in]{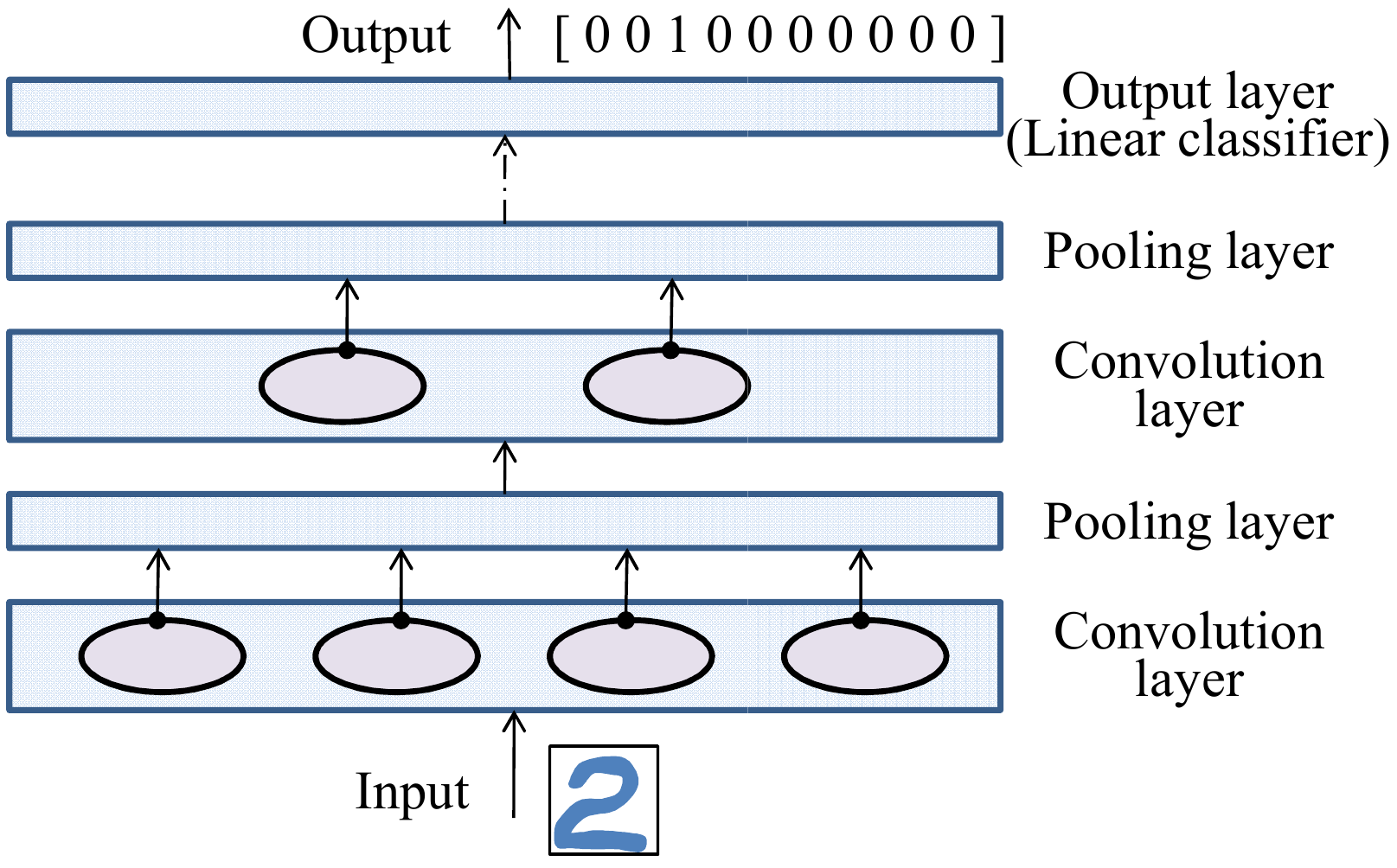}
\vspace{-0.1in}
\caption{\label{fig:cnn} \small
Convolutional neural network.  
}
\end{figure}

\subsection{Preliminary: \cnn\ for image}
\label{sec:cnn-image}

\begin{figure}
\centering
\includegraphics[width=2in]{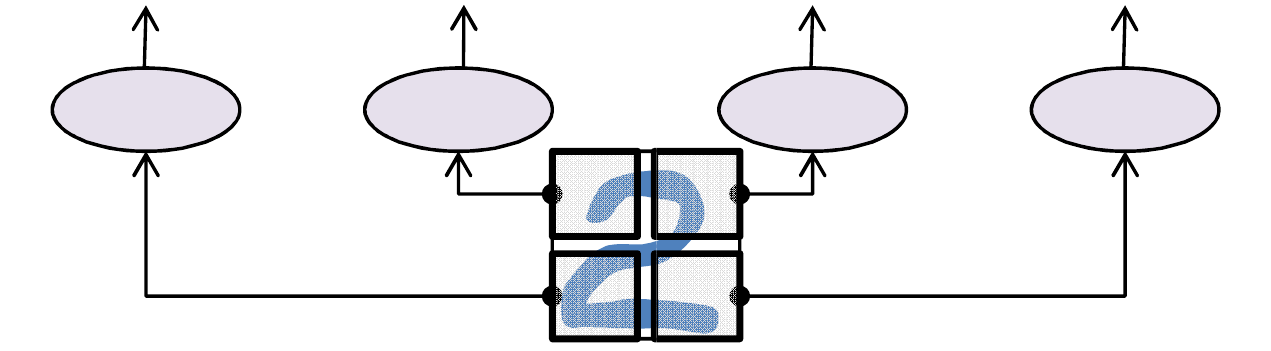}
\vspace{-0.1in}
\caption{\label{fig:imgconv} \footnotesize 
Convolution layer for image.  
Each computation unit (oval) computes a non-linear function 
$\Activ( \Wei \cdot \region_\iL(\bx) + \Bias )$ of a small region $\region_\iL(\bx)$ of input image $\bx$, 
where weight matrix $\Wei$ and bias vector $\Bias$ are shared by all the units in the same layer.
}
\end{figure}

\cnn\ is a feed-forward neural network with convolution layers interleaved with  
pooling layers, as illustrated in Figure \ref{fig:cnn}, where 
the top layer performs classification using the features generated by the 
layers below. 
A convolution layer consists of several computation units, each of which 
takes as input a {\em region vector} that represents a small region of the input image 
and applies a non-linear function to it.  
Typically, the region vector is a concatenation of pixels in the region, which 
would be, for example, 75-dimensional if the region is $5 \times 5$ and the number of {\em channels} 
is three (red, green, and blue).  
Conceptually, computation units are placed over the input image so that the entire image is 
collectively covered, as illustrated in Figure \ref{fig:imgconv}.  
The region stride (distance between the region centers) is often set to a small value such as 1 
so that regions overlap with each other, 
though the stride in Figure \ref{fig:imgconv} is set larger than the region size for illustration.  

A distinguishing feature of convolution layers is {\em weight sharing}. 
Given input $\bx$, a unit associated with the $\iL$-th region computes 
$
\Activ(\ip{\Wei}{\region_\iL(\bx)}+\Bias), 
$
where $\region_\iL(\bx)$ is a region vector representing the region of $\bx$ 
at location $\ell$, 
and $\Activ$ is a pre-defined component-wise non-linear activation function, 
(e.g., applying $\activ(x)=\max(x,0)$ to each vector component).  
The matrix of {\em weights} $\Wei$ and the vector of {\em biases} $\Bias$ are 
learned through training, and they are {\em shared} by the computation units 
in the same layer.  
This weight sharing enables learning 
useful features irrespective of their location, while preserving the 
location where the useful features appeared.  

\newcommand{\numNeurono}{m}
\newcommand{\numNeuron}{$\numNeurono$}
We regard the output of a convolution layer as an `image' so that the output of 
each computation unit is considered to be a `pixel' of \numNeuron\ channels 
where \numNeuron\ is the number of weight vectors (i.e., the number of 
rows of $\Wei$) or the number of {\em neurons}. 
In other words, {\em a convolution layer converts image regions to 
\numNeuron-dim vectors}, 
and the locations of the regions are inherited through this conversion.  
%

The output image of the convolution layer is passed to a pooling layer, 
which essentially shrinks the image by merging neighboring pixels, so that higher layers can deal with 
more abstract/global information.  
A pooling layer consists of pooling units, each of which is associated with a small region 
of the image.  
Commonly-used merging methods are average-pooling and max-pooling, 
which respectively compute the channel-wise average/maximum of each region.  



\subsection{\cnn\ for text} 
\label{sec:cnn-text}

Now we consider application of \cnn\ to text data.  
Suppose that we are given a document $\doc=(\wrd_1, \wrd_2, \ldots)$ with vocabulary 
$\voc$.  
\cnn\ requires vector representation of data that preserves internal locations (word order in this case) 
as input.  
A straightforward representation 
would be to treat each word as a pixel, treat 
$\doc$ as if it were an image of $\dsz \times 1$ pixels with $\vsz$ channels, 
and to represent each pixel (i.e., each word) as a $\vsz$-dimensional one-hot vector\footnote{
  Alternatively, one could use {\em bag-of-letter-n-gram vectors} as in \cite{Shen+etal14,Gao+etal14} 
  to cope with out-of-vocabulary words and typos.  
}. 
As a running toy example, 
suppose that vocabulary $\voc=\{$ ``don't'', ``hate'', ``I'', ``it'', ``love'' $\}$ 
and we associate the words with dimensions of vector in alphabetical order (as shown), 
and that document $\doc$=``I love it''.  
Then, we have a document vector: 
\vspace{-0.1in}
\[
  \bx=\left[~0~0~1~0~0~|~0~0~0~0~1~|~0~0~0~1~0~\right]^\top~. 
\]

\subsubsection{\scnn\ for text}
\label{sec:scnn}

As in the convolution layer for image, we 
represent each region (which each computation unit responds to) 
by a concatenation of the pixels, 
which makes $\psz\vsz$-dimensional region vectors where $\psz$ is the region size 
fixed in advance.  
For example, on the example document vector $\bx$ above, with $\psz=2$ and stride 1, 
we would have two regions 
``I love'' and ``love it'' represented by the following vectors: 
\[
\bEqsz
\region_0(\bx)=  
\left[ 
  \begin{array}{c} 0 \cr 0 \cr{\bf 1}\cr 0 \cr 0 \cr \mbox{---} \cr 0 \cr 0 \cr 0 \cr 0 \cr {\bf 1} \cr \end{array} 
\right]
\hspace{-0.1in}
  \begin{array}{c} {\rm don't}\cr{\rm hate}\cr{\rm \bf I}\cr{\rm it}\cr{\rm love}\cr   \cr  
                   {\rm don't}\cr{\rm hate}\cr{\rm I}\cr{\rm it}\cr{\rm \bf love}\cr \end{array} 
~~~~                   
\region_1(\bx)=  
\left[ 
  \begin{array}{c}  0 \cr 0 \cr 0 \cr 0 \cr {\bf 1} \cr \mbox{---} \cr 0 \cr 0 \cr 0 \cr{\bf 1}\cr 0 \cr \end{array} 
\right]
\hspace{-0.1in}
  \begin{array}{c} {\rm don't}\cr{\rm hate}\cr{\rm I}\cr{\rm it}\cr{\rm \bf love}\cr   \cr  
                   {\rm don't}\cr{\rm hate}\cr{\rm I}\cr{\rm \bf it}\cr{\rm love}\cr \end{array}       
\eEqsz
\]
The rest is the same as image; 
{\em the text region vectors are converted to feature vectors}, 
i.e., 
the convolution layer learns to {\em embed text regions} into low-dimensional vector space.   
We call a neural net with a convolution layer with this region representation 
{\em \scnn} (`seq' for keeping sequences of words) 
to distinguish it from {\em \bcnn}, described next.  

\subsubsection{\bcnn\ for text} 
\label{sec:bowconv}

A potential problem of \scnn\, 
however, is that unlike image data with 3 RGB channels, 
the number of `channels' $\vsz$ (size of vocabulary) may be very 
large (e.g., 100K), which could make each region vector $\region_\iL(\bx)$ very high-dimensional 
if the region size $\psz$ is large.  
Since the dimensionality of region vectors determines the dimensionality of weight vectors, 
having high-dimensional region vectors means more parameters to learn.  If $\psz\vsz$ is too large, 
the model becomes too complex (w.r.t. the amount of training data available) and/or training 
becomes unaffordably expensive even with efficient handling of sparse data; 
therefore, one has to lower the dimensionality by lowering 
the vocabulary size $\vsz$ and/or the region size $\psz$, which may or may not be desirable, 
depending on the nature of the task.  

An alternative we provide is to perform bag-of-word conversion to make region vectors 
$\vsz$-dimensional instead of $\psz\vsz$-dimensional; 
e.g., the example region vectors above would be converted to: 
\vspace{-0.25in}
\[
\bEqsz
\region_0(\bx)=  
\left[ 
  \begin{array}{c} 0 \cr 0 \cr{\bf 1}\cr 0 \cr{\bf 1} \end{array} 
\right]
\hspace{-0.1in}
  \begin{array}{c} {\rm don't}\cr{\rm hate}\cr{\rm \bf I}\cr{\rm it}\cr{\rm \bf love}\cr \end{array} 
~~~~                   
\region_1(\bx)=  
\left[ 
  \begin{array}{c}  0 \cr 0 \cr 0 \cr{\bf 1}\cr{\bf 1} \cr \end{array} 
\right]
\hspace{-0.1in}
  \begin{array}{c} {\rm don't}\cr{\rm hate}\cr{\rm I}\cr{\rm \bf it}\cr{\rm \bf love}\cr \end{array}
\eEqsz
\]
With this representation, we have fewer parameters to learn.  
Essentially, the expressiveness of \bconv\ (which loses word order only within small regions) 
is somewhere between 
\sconv\ and \bow\ vectors. 

\subsubsection{Pooling for text}
\begin{figure}
\centering
\includegraphics[width=2.6in]{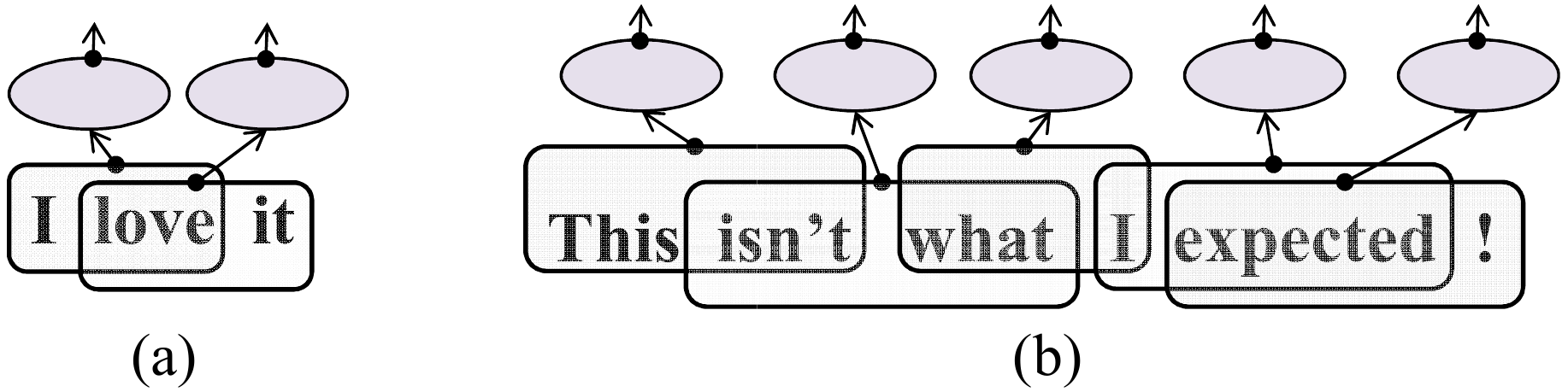}
\vspace{-0.15in}
\caption{\label{fig:txtconv} \footnotesize 
Convolution layer for variable-sized text.  
}
\end{figure}

Whereas the size of images is fixed in image applications, 
documents are naturally variable-sized, 
and therefore, with a fixed stride, the output of a convolution layer is also variable-sized 
as shown in Figure \ref{fig:txtconv}.  
Given the variable-sized output of the convolution layer, 
standard pooling for image (which uses a fixed pooling region size and a fixed stride) 
would produce variable-sized output, 
which can be passed to another convolution layer.   
To produce fixed-sized output, 
which is required by the fully-connected top layer\footnote{
  In this work, the top layer is fully-connected (i.e., each neuron responds to the entire data) 
  as in \cnn\ for image.  
  Alternatively, the top layer could be convolutional so that it can receive variable-sized input, 
  but such \cnn\ would be more complex.   
}, 
we fix the number of pooling units and dynamically
determine the pooling region size on each data point
so that the entire data is covered without overlapping.  

In the previous \cnn\ work on text, pooling is typically max-pooling 
over the entire data (i.e., one pooling unit associated with the whole text).  
The {\em dynamic $k$-max pooling} of \cite{KGB14} for sentence modeling extends it 
to take the $k$ largest values where $k$ is a function of the sentence length, 
but it is again over the entire data, and the operation is limited to max-pooling.  
Our pooling differs in that it is a natural extension of standard pooling for image, 
in which not only max-pooling but other types can be applied.  
With multiple pooling units associated with 
different regions, 
the top layer can receive locational 
information (e.g., if there are two pooling units, 
the features from the first half and last half of a document are distinguished).  
This turned out to be useful (along with average-pooling) 
on topic classification, as shown later.  


\subsection{\cnn\ vs. \bongrams}
\label{sec:cnn-vs-bow}
Traditional 
methods represent each document {\em entirely} with 
one \bongram\ vector 
and then apply a classifier model such as SVM. 
However, 
since 
high-order $n$-grams are 
susceptible to data sparsity, use of a large $n$ such as 20 is not only 
infeasible but also ineffective.  
Also 
note that a \bongram\ represents each $n$-gram by a one-hot vector 
and ignores the fact that some $n$-grams share constituent words.  
By contrast, \cnn\  
internally learns {\em embedding of text regions} (given the consituent words as input) 
{\em useful for the intended task}.  
Consequently, 
a large $n$ such as 20 can be used especially with the \bconv\ layer, 
which turned out to be useful on topic classification.
A neuron 
trained to assign a large value to, e.g., ``I love'' (and a small value to ``I hate'') 
is likely to assign a large value to ``we love'' (and a small value to ``we hate'') 
as well, {\em even though ``we love'' was never seen during training}.  
We will confirm these points empirically later.  

\subsection{Extension: parallel \cnn}

\begin{figure}
\centering
\includegraphics[width=3in]{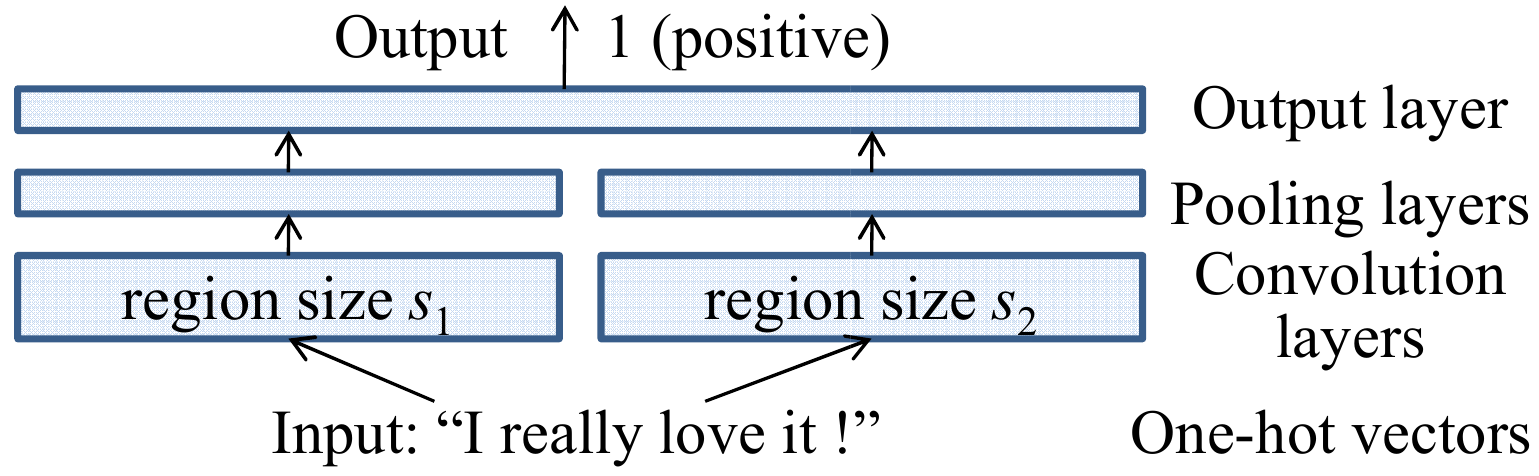}
\caption{\label{fig:multilaycnn} \footnotesize 
CNN with two convolution layers in parallel.  
}
\end{figure}

We have described \cnn\ with the simplest network architecture that has one pair of 
convolution and pooling layers.  While this can be extended in several ways (e.g., 
with deeper layers), in our experiments, 
we explored {\em parallel \cnn}, which has 
two or more convolution layers in parallel\footnote{
  Similar architectures have been used for image.  
  \newcite{Kim14} used it for text, but it was on top of a word vector conversion layer. 
}, 
as illustrated in Figure \ref{fig:multilaycnn}. 
The idea is to learn multiple types of embedding of small text regions so that they can complement 
each other to improve model accuracy.  
In this architecture, 
multiple convolution-pooling pairs with different region sizes (and possibly 
different region vector representations) are given one-hot vectors as input and produce 
feature vectors for each region; the top layer takes the concatenation of the produced 
feature vectors as input.  

\section{Experiments}
\label{sec:experiment} 

We experimented with 
\cnn\ on two tasks, topic classification and sentiment classification. 
%
Detailed information for reproducing the results 
is available on the internet along with our code.  

\subsection{\cnn}

We fixed the activation function to 
rectifier {\small $\activ(x)=\max(x,0)$} and minimized square loss with $L_2$ 
regularization by 
{\em stochastic gradient descent} (SGD).  
We only used the 30K words that appeared most frequently in the training set; 
thus, for example, 
in \scnn\ with region size 3, a region vector is 90K dimensional.  
Out-of-vocabulary words were represented by a zero vector.  
On \bcnn, to speed up computation, 
we used {\em variable region stride} 
so that a larger stride was taken where repetition\footnote{
  For example, if we slide a window of size 3 over ``* * foo * *''
  where ``*'' is out of vocabulary, 
  a bag of ``foo'' will be repeated three times with stride fixed to 1.   
} 
of the same region vectors
can be avoided by doing so.  
Padding\footnote{
  As is commonly done, 
  to the beginning and the end of each document, special words that are treated as unknown words 
  (and converted to zero vectors instead of one-hot vectors) were added as `padding'.
  The purpose is to equally treat the words at the edge and words in the middle.  
}
size was fixed to $p-1$ where $p$ is the region size.  

We used two techniques commonly used with \cnn\ on image, which typically 
led to small performance improvements.  
One is {\em dropout} \cite{dropout12} optionally applied to the input to the top layer.  
The other is {\em response normalization} as in \cite{imagenetNips12}, 
which in our case scales the output of the pooling layer 
$\bz$ at each location by multiplying $(1 + |\bz|^2)^{-1/2}$.  

\subsection{Baseline methods} 
For comparison, we tested SVM
with the linear kernel and 
fully-connected neural networks
(see e.g., \newcite{Bishop95})
with \bongram\ vectors as input.  
To experiment with fully-connected neural nets, 
as in \cnn, we minimized square loss 
with $L_2$ regularization 
and optional dropout
by SGD, and activation was fixed 
to rectifier.  
To generate \bongram\ vectors, 
on topic classification, 
we first set each component to $\log(x+1)$ 
where $x$ is the word frequency in the document and then scaled them to unit vectors, 
which we found always 
improved performance over raw frequency.  
On sentiment classification, as is often done, we generated binary vectors and scaled them 
to unit vectors. 
We tested three types of \bongram: 
\bowone\ with $n \in \{1\}$, \bowtwo\ with $n \in \{1,2\}$, and \bowthree\ with $n \in \{1,2,3\}$; 
that is, \bowone\ is the traditional \bow\ vectors, and 
with \bowthree, each component of the vectors corresponds to either uni-gram, bi-gram, 
or tri-gram of words.  


We used SVMlight\footnote{
  \url{http://svmlight.joachims.org/}
}
for the SVM experiments.  
\tightpara{\nbw}
We also tested \nbw, which first appeared (but without performance report\footnote{
  WM12 instead reported the performance of an ensemble of NB and SVM as it performed better.  
}
) as NBSVM in WM12 \cite{WM12} and later with a small modification 
produced 
performance that exceeds state-of-the-art supervised methods 
on IMDB (which we experimented with) 
in MMRB14 \cite{MMRB14}.  
We experimented with the MMRB14 version, which 
generates binary \bongram\ vectors, multiplies the component 
for each $n$-gram $f_i$ with $\log(P(f_i|Y=1)/P(f_i|Y=-1))$
({\em NB-weight}) where the probabilities are estimated using the training data, 
and does logistic regression training.  
We used MMRB14's software\footnote{
  \url{https://github.com/mesnilgr/nbsvm}
} with a modification so that the regularization parameter
can be tuned on development data.  

\subsection{Model selection}
\label{sec:protocol}
For all the methods, 
the hyper-parameters such as net configurations and 
regularization parameters were chosen 
based on the performance on the development data (held-out portion of the training data), 
and using the chosen hyper-parameters, the models were re-trained using all the training 
data.   


\subsection{Data, tasks, and data preprocessing}
\tightpara{IMDB: movie reviews}
The IMDB dataset 
\cite{MDPHNP11}
is a benchmark dataset for sentiment classification.  
The task is to determine if the movie reviews are positive or negative.  
Both the training and test sets consist of 25K reviews.  
For preprocessing, we tokenized the text so that emoticons such as ``:-)'' 
are treated as tokens and converted all the characters to lower case.  

\tightpara{\Elec: electronics product reviews}
\Elec\ consists of electronic product reviews.  It is  
part of a large Amazon review dataset
\cite{ML13}.  
We chose electronics as it seemed to be very different from movies.    
Following the generation of IMDB \cite{MDPHNP11}, we chose the training set and 
the test set so that one half of each set consists of positive reviews 
and the other half is negative, regarding rating 1 and 2 as negative and 4 and 5 
as positive, and that the reviewed products are disjoint between the training set 
and test set.  Note that to extract text from the original data, 
we {\em only} used the {\em text section}, and we did {\em not} use the {\em summary section}.  
This way, we obtained a test set of 25K reviews (same as IMDB) 
and training sets of various sizes.  
The training and test sets are available on the internet\footnote{
  \url{riejohnson.com/cnn_data.html}
}. 
Data preprocessing was the same as IMDB.

\tightpara{RCV1: topic categorization} 
RCV1 
is a corpus of Reuters news articles 
as described in LYRL04 \cite{rcv1}.  
RCV1 has 103 topic categories 
in a hierarchy, 
and one document may 
be associated with more than one topic.  
Performance on this task (multi-label categorization) 
is known to be sensitive to thresholding strategies, 
which are algorithms additional to the models we would like to test.  
Therefore, we also experimented with single-label categorization 
to assign one of 55 second-level topics to each document 
to directly evaluate models.  
For this task, 
we used the documents from a one-month period 
as the test set and generated various sizes of 
training sets from the documents with {\em earlier} dates.  
Data sizes are shown in Table \ref{tab:rcv-data}.  
As in 
LYRL04, we used the concatenation of the headline and text elements. 
Data preprocessing was the same as IMDB except that we used the stopword list 
provided by LYRL04 and regarded numbers as stopwords.  

\begin{table}
\begin{center}
\begin{footnotesize}
\begin{tabular}{|c|c|r|r|r|} 
\hline
   & label          &\multicolumn{1}{|c|}{\#train}&\multicolumn{1}{|c|}{\#test}& \#class \\
\hline
Table \ref{tab:all}  & single  & 15,564 & 49,838 & 55 \\
\hline
Fig. \ref{fig:size}  & single  & varies &  
                                          49,838 & 55 \\
\hline                                                                
Table \ref{tab:rcv-multi}   & multi   & 23,149 & 781,265 & 103 \\
\hline
\end{tabular}
\end{footnotesize}
\vspace{-0.1in}
\caption{ \label{tab:rcv-data} \small 
RCV1 data summary.  
}
\end{center}
\end{table}

\subsection{Performance results}
Table \ref{tab:all} shows the error rates of \cnn\ in comparison with 
the baseline methods. 
The first thing to note is that on all the datasets, 
the best-performing \cnn\ outperforms the baseline methods, 
which demonstrates the effectiveness of our approach.  
  
To look into the details, 
let us first focus on \cnn\ with one convolution layer 
(\scnnpfx- and \bcnn\ in the table).  
On sentiment classification (IMDB and \Elec), 
the configuration chosen by model selection 
was: region size 3, stride 1, 1000 weight vectors, and max-pooling 
with one pooling unit, for both types of \cnn; 
\scnn\ outperforms \bcnn, as well as all the baseline methods except for one.  
Note that with a small region size and max-pooling, 
if a review contains a short phrase that conveys strong sentiment 
(e.g., ``A great movie!''), the review could receive a high score irrespective of the rest of the review. 
It is sensible that this type of configuration is effective on sentiment classification.  

By contrast, on topic categorization (RCV1), 
the configuration chosen for \bcnn\ by model selection was: region size 20, variable-stride$\ge$2, 
average-pooling with 10 pooling units, and 1000 weight vectors, 
which is very different from sentiment classification.  
This is presumably because 
on topic classification, a larger context would be more predictive 
than short fragments ($\rightarrow$ larger region size), 
the entire document matters ($\rightarrow$ the effectiveness of average-pooling), 
and the location of predictive text also matters ($\rightarrow$ multiple pooling units). 
The last point may be because news documents tend to have crucial sentences 
(as well as the headline) at the beginning.  
On this task, while both \scnnpfx\ and \bcnn\ outperform the baseline methods, \bcnn\ outperforms \scnn,  
which indicates that in this setting 
the merit of having fewer parameters is larger than the benefit of keeping word order 
in each region.  
 
Now we turn to parallel \cnn. 
On IMDB, \sscnn, which has two seq-convolution layers 
(region size 2 and 3; 1000 neurons each; followed by one unit of max-pooling each), 
outperforms \scnn.  With more neurons (3000 neurons each; Table \ref{tab:imdb-prev}) it 
further exceeds 
the best-performing baseline, which is also the best previous supervised result.    
We presume the effectiveness of \sscnn\ indicates that the length of predictive text regions 
is variable.  

The best performance 7.67 on IMDB was obtained by `\ssbcnn', 
equipped with three layers in parallel: 
two seq-convolution layers (1000 neurons each) as in \sscnn\ above and  
one layer (20 neurons) that {\em regards the entire document as one region} and 
represents the region (document) by a \bongram\ vector (\bowthree)
as input to the computation unit; 
in particular, we generated \bowthree\ vectors by multiplying the NB-weights with binary vectors, 
motivated by the good performance of \nbw.  
This third layer is a \bow-convolution layer\footnote{
  It can also be regarded as a fully-connected layer that takes \bowthree\ vectors as input. 
}
with one region of variable size that takes 
one-hot vectors with $n$-gram vocabulary as input to learn document embedding.  
The \ssbcnn\ for \Elec\ in the table is the same except that the regions sizes of seq-convolution layers are 
3 and 4.  On both datasets, performance is improved over \sscnn.
The results 
suggest that what can be learned through these three layers are distinct enough 
to complement each other.  
The effectiveness of the third layer indicates that not only short word sequences 
but also global context in a large window may be useful on this task; thus, inclusion of a 
\bconv\ layer with $n$-gram vocabulary with a large fixed region size might be even more effective, 
providing more focused context, 
but we did not pursue it in this work.  


\tightpara{Baseline methods} 
Comparing the baseline methods with each other, on sentiment classification, 
reducing the vocabulary to the most frequent $n$-grams notably hurt 
performance (also observed on \nbw\ and NN)
even though some reduction is a common practice. 
Error rates were clearly improved by addition of bi- and tri-grams. 
By contrast, on topic categorization, 
bi-grams only slightly improved accuracy, and reduction of vocabulary did not 
hurt performance.  
\nbw\ is very strong on IMDB and poor on RCV1; its effectiveness appears to be 
data-dependent, 
as also observed by WM12. 

\begin{table}
\begin{center}
\begin{small}
\begin{tabular}{|l|r|r|r|} 
\hline
methods        & IMDB  & Elec  &  RCV1 \\  %
\hline
SVM \bowthree\ (30K)& 10.14 &  9.16 & 10.68 \\ %
\hline
SVM \bowone\ (all)  &    11.36 &    11.71 &     10.76 \\ 
SVM \bowtwo\ (all)  &     9.74 &     9.05 &     10.59 \\ 
SVM \bowthree\ (all)&     9.42 &     8.71 &     10.69 \\
\hline
NN \bowthree\ (all) &     9.17 &     8.48 &     10.67 \\ %
\hline
\nbw\ \bowthree\ (all)&     8.13 &     8.11 &     13.97 \\ 
\hline \hline
\bcnn\      &     8.66 &     8.39 &{\bf 9.33}\\ 
\scnn\      &     8.39 &     7.64 &     9.96 \\ 
\hline
\sscnn\     &     8.04 &     7.48 &\multicolumn{1}{|c|}{--}\\
\ssbcnn\    &{\bf 7.67}&{\bf 7.14}&\multicolumn{1}{|c|}{--}\\
\hline
\end{tabular}
\end{small}
\vspace{-0.1in}
\caption{ \label{tab:all} \small
Error rate (\%) 
comparison with bag-of-$n$-gram-based methods.
Sentiment classification on IMDB and \Elec\
(25K training documents) 
and 55-way topic categorization on RCV1 (16K training documents).  
`(30K)' indicates that the 30K most frequent $n$-grams were used, 
and `(all)' indicates that all the $n$-grams (up to 5M) were used.  
\cnn\ used the 30K most frequent words.  
}
\end{center}
\end{table}

\begin{table}
\begin{center}
\begin{small}
\begin{tabular}{|l|r|c|} 
\hline
SVM \bowtwo\ [WM12]    & 10.84 & -- \\
WRRBM+\bow\ [DAL12]    & 10.77 & -- \\
NB+SVM \bowtwo\ [WM12] & 8.78  & ensemble\\
\nbw\ \bowthree\ [MMRB14]& 8.13  & -- \\
\hline
Paragraph vectors [LM14] & 7.46 & unlabeled data\\
\hline  
\sscnn\ (3K$\times$2) [Ours]  &     7.94  & -- \\   
\ssbcnn\ [Ours] &{\bf 7.67} & -- \\
\hline
\end{tabular}
\end{small}
\end{center}
\vspace{-0.2in}
\caption{ \label{tab:imdb-prev} \small
Error rate (\%) comparison with previous best methods 
on IMDB. 
}
\end{table}

\tightpara{Comparison with state-of-the-art results} 
As shown in Table \ref{tab:imdb-prev}, 
the previous best supervised result on IMDB is 8.13 by \nbw\ with \bowthree\ (MMRB14), 
and our best error rate 7.67 is better by nearly 0.5\%.  
\cite{LM14} reports 7.46 with the semi-supervised method that learns low-dimensional 
vector representations of documents from unlabeled data.  
Their result is not directly comparable with our supervised results due to use of additional resource.  
Nevertheless, our best result rivals their result.  


We tested \bcnn\ on the multi-label topic categorization task on RCV1
to compare with LYRL04.  
%
We used the same thresholding strategy as LYRL04. 
As shown in Table \ref{tab:rcv-multi}, 
\bcnn\ outperforms LYRL04's best results 
even though our data preprocessing is much simpler (no stemming and no tf-idf weighting).  

\begin{table}
\begin{center}
\begin{small}
\begin{tabular}{|c|c|c|} 
\hline
models                         & micro-F & macro-F \\
\hline
LYRL04's best SVM   & 81.6   & 60.7    \\
\hline
\bcnn\                          &{\bf 84.0}&{\bf 64.8}\\ 
\hline
\end{tabular}
\end{small}
\vspace{-0.1in}
\caption{ \label{tab:rcv-multi} \small
RCV1 micro-averaged 
and macro-averaged F-measure results 
on multi-label task with LYRL04 split.   
}
\end{center}
\end{table}

\tightpara{Previous \cnn}
We focus on the sentence classification studies due to its relation to text categorization.  
\newcite{Kim14} studied fine-tuning of pre-trained word vectors to produce input to 
parallel \cnn.  
He reported that performance was poor 
when word vectors were trained as part of \cnn\ training (i.e., no additional method/corpus).  
On our tasks, 
we were also unable to outperform the baselines with this type of model. 
%
Also, with our approach, a system is simpler with one fewer layer 
-- no need to tune the dimensionality of word vectors or meta-parameters for word vector learning.  

\begin{figure}[t]
\centering
\includegraphics[width=3.2in]{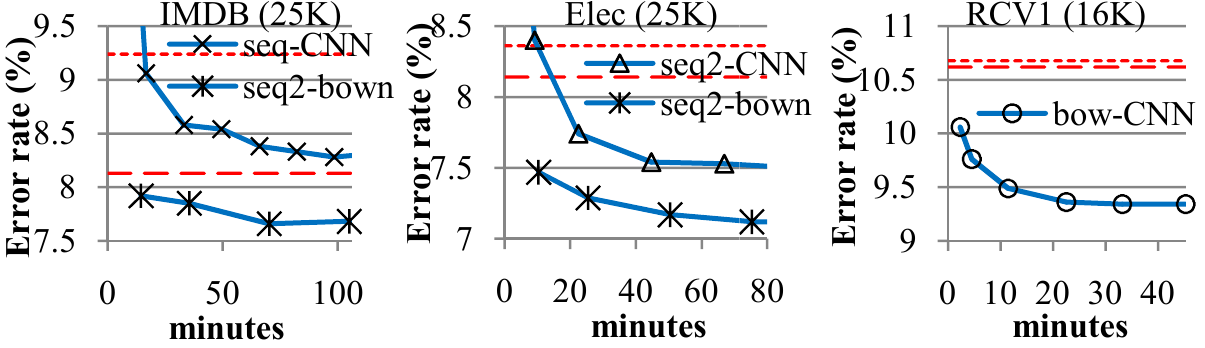}
\vspace{-0.3in}
\caption{\label{fig:time} \small 
Training time (minutes) on 
Tesla K20. 
The horizontal lines are 
the best-performing baselines.  
}
\end{figure}

\newcite{KGB14} proposed complex modifications of \cnn\ 
for sentence modeling.  Notably, given word vectors $\in \mathbb{R}^d$, 
their convolution with \numNeuron\ feature maps produces for each region a matrix $\in \mathbb{R}^{d \times \numNeurono}$ 
(instead of a vector $\in \mathbb{R}^\numNeurono$ as in standard \cnn). 
Using the provided code, we found that their model is too resource-demanding 
for our tasks.  
On IMDB and \Elec\footnote{
  We could not train adequate models on RCV1 on either Tesla K20 or M2070 
  due to memory shortage. 
}
the best error rates we obtained
by training with various configurations that fit in memory 
for 24 hours each on GPU (cf. Fig \ref{fig:time}) 
were 10.13 and 9.37, respectively, 
which is no better than SVM \bowtwo.  
Since excellent performances were reported on short sentence classification, 
we presume that their model is optimized for short sentences, but not for text categorization in general.   



\begin{figure}[t]
\centering
\includegraphics[width=3.2in]{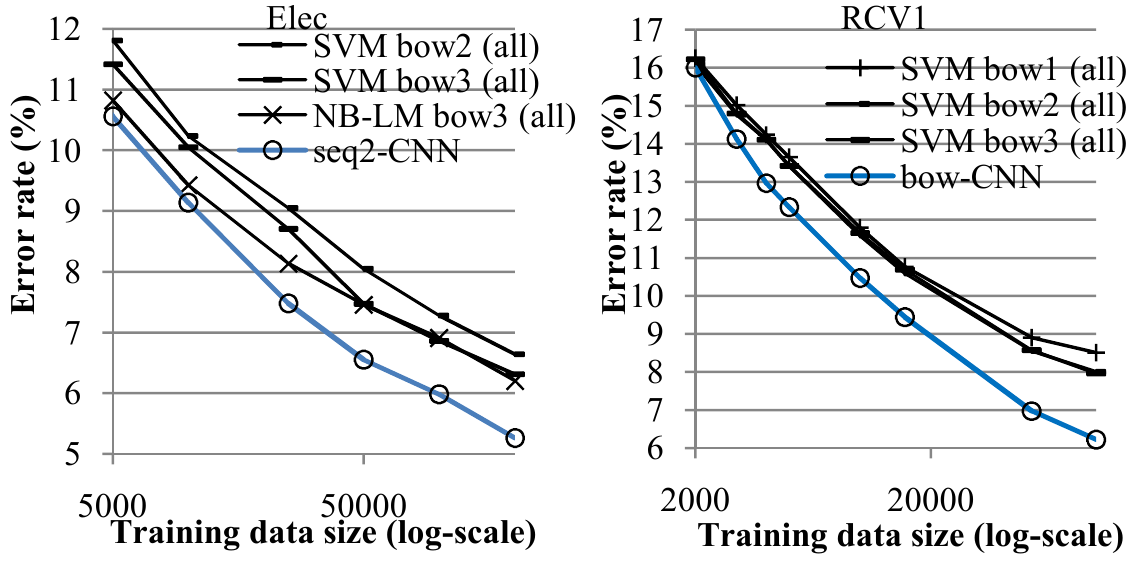}
\vspace{-0.3in}
\caption{\label{fig:size} \small
Error rate in relation to training data size. 
For readability, only representative methods are shown. 
}
\end{figure}
\tightpara{Performance dependency}
%
\cnn\ training is known to be expensive, compared with, e.g., linear models -- 
linear SVM with \bowthree\ on IMDB only 
takes 9 minutes using SVMlight (single-core) on a high-end Intel CPU.  
Nevertheless, with our code on GPU, \cnn\ training only takes minutes (to a few hours)
on these datasets shown in Figure \ref{fig:time}.  

Finally, the results with training sets of various sizes on \Elec\ and RCV1 
are shown in Figure \ref{fig:size}.


\subsection{Why is \cnn\ effective?} 
\label{sec:examples}
In this section we explain the effectiveness of \cnn\ through 
looking into what it learns from training.  

First, for comparison,
we show the $n$-grams that SVM with \bowthree\ found to be the most predictive; i.e., 
the following $n$-grams were assigned the 10 largest weights by SVM with binary features on \Elec\ 
for the negative and positive class, respectively: 
\vspace{-0.1in}
\begin{itemize} \itemsep1pt \parskip0pt \parsep0pt
\item 
{\small
  poor, useless, returned, not worth, return, worse, disappointed, terrible, worst, horrible  
}
\item
{\small
  great, excellent, perfect, love, easy, amazing, awesome, no problems, perfectly, beat
}  
\end{itemize}
\vspace{-0.05in}
Note that, even though SVM was also given bi- and tri-grams, 
the top 10 features chosen by SVM with binary features are mostly uni-grams; 
furthermore, the top 100 features (50 for each class) include 28 bi-grams but only four tri-grams. 
This means that, with the given size of training data, SVM still heavily counts on 
uni-grams, which could be ambiguous, and cannot fully take advantage of higher-order 
$n$-grams.
By contrast, 
NB-weights tend to promote $n$-grams with a larger $n$; the 100 features that were assigned 
the largest NB-weights are 7 uni-, 33 bi-, and 60 tri-grams.  
However, as seen above, NB-weights do not always lead to the best performance.  

In Table \ref{tab:bv}, 
we show some of text regions learned by \scnn\ to be predictive on \Elec.  
This net has one convolution layer with region size 3 and 1000 neurons; 
thus, embedding by the convolution layer produces a 1000-dim vector 
for each region, which (after pooling) serves as features in the top layer
where weights are assigned to the 1000 vector components.  
In the table, N$i$/P$i$ indicates the component that received 
the $i$-th highest weight in the top layer 
for the negative/positive class, respectively. The table shows the text regions 
(in the training set) whose embedded vectors have a large value
in the corresponding component, i.e., predictive text regions.  

Note that the embedded vectors for the text regions listed in the same row 
are close to each other as they have a large value in the same component. 
That is, Table \ref{tab:bv} also shows that
the {\em proximity of the embedded vectors} tends to reflect the {\em proximity  
in terms of the relations to the target classes} (positive/negative sentiment). 
This is the effect of embedding, which helps classification by the top layer.  


\begin{table}
\begin{center}
\begin{footnotesize}
\begin{tabular}{|c|p{2.5in}|} 
\hline
{\small N1} & {\small completely useless ., return policy .} \\
{\small N2} & {\small it won't even, but doesn't work} \\
{\small N3} & {\small product is defective, very disappointing !} \\
{\small N4} & {\small is totally unacceptable, is so bad} \\
{\small N5} & {\small was very poor, it has failed} \\
\hline
{\small P1} & {\small works perfectly !, love this product} \\
{\small P2} & {\small very pleased !, super easy to, i am pleased} \\
{\small P3} & {\small 'm so happy, it works perfect, is awesome !} \\
{\small P4} & {\small highly recommend it, highly recommended !} \\  
{\small P5} & {\small am extremely satisfied, is super fast} \\
\hline
\end{tabular}
\end{footnotesize}
\vspace{-0.1in}
\caption{ \label{tab:bv} \small
Examples of predictive text regions in the training set. 
}
\end{center}
\end{table}

With the \bongram\ representation, 
only the $n$-grams that 
appear in the training data can participate in prediction. 
By contrast, 
one strength of \cnn\ is that $n$-grams (or text regions of size $n$) 
{\em can contribute to accurate prediction even if they did not appear in the training data}, 
as long as (some of) their constituent words did, 
because input of embedding is the constituent words of the region.   
%
\begin{table}
\begin{center}
\begin{footnotesize}
\begin{tabular}{|p{2.9in}|} 
\hline
{\small 
were unacceptably bad, is abysmally bad,  
were universally poor, 
was hugely disappointed, was enormously disappointed, 
is monumentally frustrating,  are endlessly frustrating
}\\
\hline
{\small 
best concept ever, best ideas ever, best hub ever, 
am wholly satisfied, 
am entirely satisfied, am incredicbly satisfied, 
'm overall impressed, am awfully pleased, 
am exceptionally pleased, 'm entirely happy, 
are acoustically good, is blindingly fast, 
}\\
\hline
\end{tabular}
\end{footnotesize}
\vspace{-0.1in}
\caption{ \label{tab:bv-tstonly} \small
Examples of text regions that contribute to prediction. 
They are from the {\em test set}, and they 
did {\em not} appear in the training set, either entirely or partially as bi-grams.  
}
\end{center}
\end{table}
%
To see this point, in Table \ref{tab:bv-tstonly} we show the text regions from the {\em test set}, 
which {\em did not appear in the training data}, either entirely or partially as bi-grams, 
and yet whose embedded features have large values in the heavily-weighted (predictive) component 
thus contributing to the prediction.  
There are many more of these, and we only show a small part of them that fit certain patterns.  
One noticeable pattern is (be-verb, adverb, sentiment adjective) such as 
``am entirely satisfied'' and ``'m overall impressed''.  
These adjectives alone could be ambiguous as they may be negated.  To know that the 
writer is indeed ``satisfied'', we need to see the sequence ``am satisfied'', 
but the insertion of adverb such as ``entirely''  
is very common.  ``best X ever' is another pattern that a discriminating pair of words are not adjacent to 
each other.  These patterns require tri-grams for disambiguation, and 
\scnn\ successfully makes use of them even though the exact tri-grams were not seen during training, 
as a result of learning, e.g., ``am X satisfied'' with non-negative X 
(e.g., ``am very satisfied'', ``am so satisfied'') 
to be predictive of the positive class through training. 
%
That is, 
\cnn\ can effectively use word order when \bongram-based approaches fail.  

\section{Conclusion}
\label{sec:conclude}
This paper showed that \cnn\ provides an alternative mechanism for effective use of word order 
for text categorization through direct embedding of small text regions, 
different from the traditional \bongram\ approach or word-vector \cnn.  
With the parallel \cnn\ framework, several types of embedding 
can be learned and combined so that they can complement each other 
for higher accuracy.  
State-of-the-art performances on sentiment classification and topic classification were 
achieved using this approach. 

\section*{Acknowledgements}
We thank the anonymous reviewers for useful suggestions.  
The second author was supported by NSF IIS-1250985 and NSF IIS-1407939.  

\bibliographystyle{naaclhlt2015}
\bibliography{cnn-text-naacl-arxiv}

\end{document}